\documentclass{article}

\usepackage[preprint]{neurips_2024}

\usepackage[utf8]{inputenc} 
\usepackage[T1]{fontenc}    
\usepackage{url}            
\usepackage{booktabs}       
\usepackage{amsfonts}       
\usepackage{nicefrac}       
\usepackage{microtype}      
\usepackage{xcolor}         

\usepackage{amsmath}
\usepackage{amssymb}
\usepackage{mathtools}
\usepackage{amsthm}

\usepackage{caption}
\usepackage{subcaption}

\usepackage{multirow}
\usepackage{comment} 
\usepackage{enumitem}

\usepackage{times}
\usepackage{epsfig}
\usepackage{graphicx}
\usepackage{wrapfig}

\usepackage[bb=dsserif]{mathalpha}
\usepackage{bm}

\usepackage{booktabs,colortbl,tabularx}
\usepackage{pifont}%

\definecolor{Gray}{gray}{0.9}
\definecolor{citecolor}{HTML}{2980b9}
\definecolor{linkcolor}{HTML}{c0392b}

\usepackage[pagebackref=true,breaklinks=true,colorlinks,bookmarks=false,citecolor=citecolor,linkcolor=linkcolor]{hyperref}

\usepackage[capitalize,noabbrev]{cleveref}

\definecolor{battleshipgrey}{rgb}{0.52, 0.52, 0.51}

\title{Scaling Diffusion Mamba with Bidirectional SSMs for Efficient Image and Video Generation}

\author{%
  Shentong Mo\\
  Carnegie Mellon University
  \And
  Yapeng Tian \\
  University of Texas at Dallas 
}

\begin{document}

\maketitle

\begin{abstract}

In recent developments, the Mamba architecture, known for its selective state space approach, has shown potential in the efficient modeling of long sequences.
However, its application in image generation remains underexplored. Traditional diffusion transformers (DiT), which utilize self-attention blocks, are effective but their computational complexity scales quadratically with the input length, limiting their use for high-resolution images. 
To address this challenge, we introduce a novel diffusion architecture, Diffusion Mamba (DiM), which foregoes traditional attention mechanisms in favor of a scalable alternative. 
By harnessing the inherent efficiency of the Mamba architecture, DiM achieves rapid inference times and reduced computational load, maintaining linear complexity with respect to sequence length. 
Our architecture not only scales effectively but also outperforms existing diffusion transformers in both image and video generation tasks. 
The results affirm the scalability and efficiency of DiM, establishing a new benchmark for image and video generation techniques. 
This work advances the field of generative models and paves the way for further applications of scalable architectures.

\end{abstract}

\section{Introduction}

The quest for efficient and scalable image generation models~\cite{ho2020denoising,song2021scorebased,song2021denoisingdi} is a central pursuit in the field of machine learning, particularly in generative modeling.
Recently, we have seen significant strides in this area, with diffusion models~\cite{ho2020denoising,song2021scorebased} and self-attention transformers~\cite{Peebles2022DiT,bao2022all,bao2023transformer} at the forefront of innovation.
These models have demonstrated remarkable capabilities in capturing intricate details and producing high-quality images. 
However, as we push the boundaries towards generating higher-resolution images, the computational complexity of these models, especially their quadratic scaling with respect to the input length, emerges as a formidable barrier to scalability and efficiency.

Concurrently, the Mamba architecture~\cite{gu2023mamba}, known for its selective state space approach, has shown great promise in modeling long sequences efficiently. 
The architecture’s design allows it to handle complex dependencies with significantly reduced computational requirements, making it a potential candidate for addressing the challenges in image generation. 
However, the direct application of Mamba principles to image generation has not been thoroughly explored, leaving a gap in the current landscape of generative models.

To address this gap, we introduce a novel architecture, the Diffusion Mamba (DiM), which harmonizes the efficiency of the Mamba architecture with the generative prowess of diffusion models. 
DiM departs from conventional reliance on attention mechanisms, opting instead for a structure that supports fast inference and boasts lower floating-point operations per second (FLOPs) while maintaining a linear complexity with sequence length. 
This innovative approach not only addresses the computational inefficiencies of previous models but also ensures scalability to high-resolution image generation without compromising on image quality.

\begin{figure*}[t]
\centering
\includegraphics[width=\linewidth]{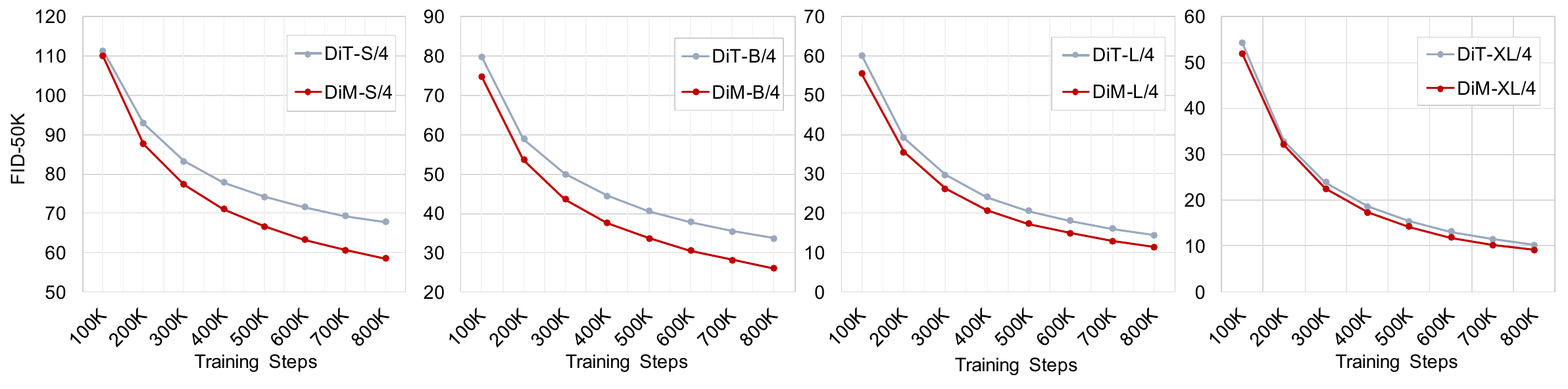}
\vspace{-1.0em}
\caption{{\bf Comparison with DiT~\cite{Peebles2022DiT} on FID-50k across all model sizes.} Our DiM models achieve better results across all training steps for all model sizes.
}
\label{fig: title_image}
\vspace{-1.0em}
\end{figure*}

In this work, we also present a comprehensive analysis of the limitations inherent in existing diffusion models and self-attention transformers, particularly their computational inefficiency at scale. 
Furthermore, we detail the development and implementation of DiM, elucidating its architectural novelties and the theoretical underpinnings that facilitate its enhanced performance and scalability. 
We benchmark DiM against state-of-the-art models, including the diffusion transformer (DiT), across multiple resolutions. 
Our results affirm DiM's superior performance and lower computational footprint, setting new precedents for efficient, scalable image generation and video generation.

Our main contributions can be summarized as follows:
\begin{itemize}
    \item We introduce the Diffusion Mamba architecture, namely DiM, a novel approach that integrates the computational efficiency of the Mamba state space models with the generative capabilities of diffusion processes.
    \item Our DiM not only adapts well to various operational scales but also maintains high efficiency, significantly lowering the Gflops required compared to previous methods.
    \item Extensive experimental results validate the effectiveness of the DiM architecture across multiple standard datasets including ImageNet for images and UCF-101 for videos.
\end{itemize}

\section{Related Work}

\noindent\textbf{State Space Models.}
State space models~\cite{gu2023mamba,fu2023hungry,gu2022efficiently} have seen a resurgence in machine learning, particularly in handling sequences efficiently. 
Current works like S4~\cite{gu2022efficiently} and Mamba~\cite{gu2023mamba} by Gu et al. have demonstrated that SSMs can efficiently process long sequences with fewer parameters and reduced computational overhead compared to traditional recurrent neural networks. 
Our work extends these principles to the image and video generation domains, leveraging the efficiency of state space models to enhance the performance and scalability of generative tasks.

\noindent\textbf{Diffusion Models.}
Diffusion models have emerged as a powerful class of generative models, capable of synthesizing high-quality images~\cite{ho2022imagen}, restoring images, and generating speech~\cite{kong2021diffwave}. Foundational models like the denoising diffusion probabilistic models (DDPMs) introduced by Ho et al.~\cite{ho2020denoising} and extended in various forms by Song et al.~\cite{song2021scorebased,song2021denoisingdi} have set benchmarks in several domains. 
For instance, Photorealistic Image Generation by Saharia et al.~\cite{saharia2022photorealistic} and Image Restoration by Saharia et al.~\cite{saharia2021image} showcase the versatility of diffusion models in handling diverse and complex tasks. 
Our DiM architecture integrates these diffusion principles with the computational efficiency of state space models to improve performance in image and video generation.

\noindent\textbf{Diffusion Transformers.}
Diffusion Transformers have been successful in generating not only high-fidelity images~\cite{Peebles2022DiT,bao2022all,bao2023transformer,xie2023difffit} but also complex 3D structures~\cite{mo2023dit3d,mo2023fastdit3d}. 
The Diffusion Transformer (DiT) proposed by Peebles et al.\cite{Peebles2022DiT} introduces a ViT~\cite{Dosovitskiy2021vit}-like method to learn the denoising process on latent patches extracted via a pre-trained variational autoencoder in Stable Diffusion~\cite{Rombach2022highresolution}, showing significant improvements in image quality. 
Extensions of this concept, such as U-ViT~\cite{bao2022all} and UniDiffuser~\cite{bao2023transformer} by Bao et al., illustrate the adaptability of diffusion transformers across various modalities by manipulating conditions and integrating multi-modal inputs within a unified framework. 
In this work, we explore the scalability and efficiency of these models, providing a novel approach to managing the generative process with an enhanced focus on reducing computational demands.

\begin{figure*}[t]
\centering
\includegraphics[width=0.99\linewidth]{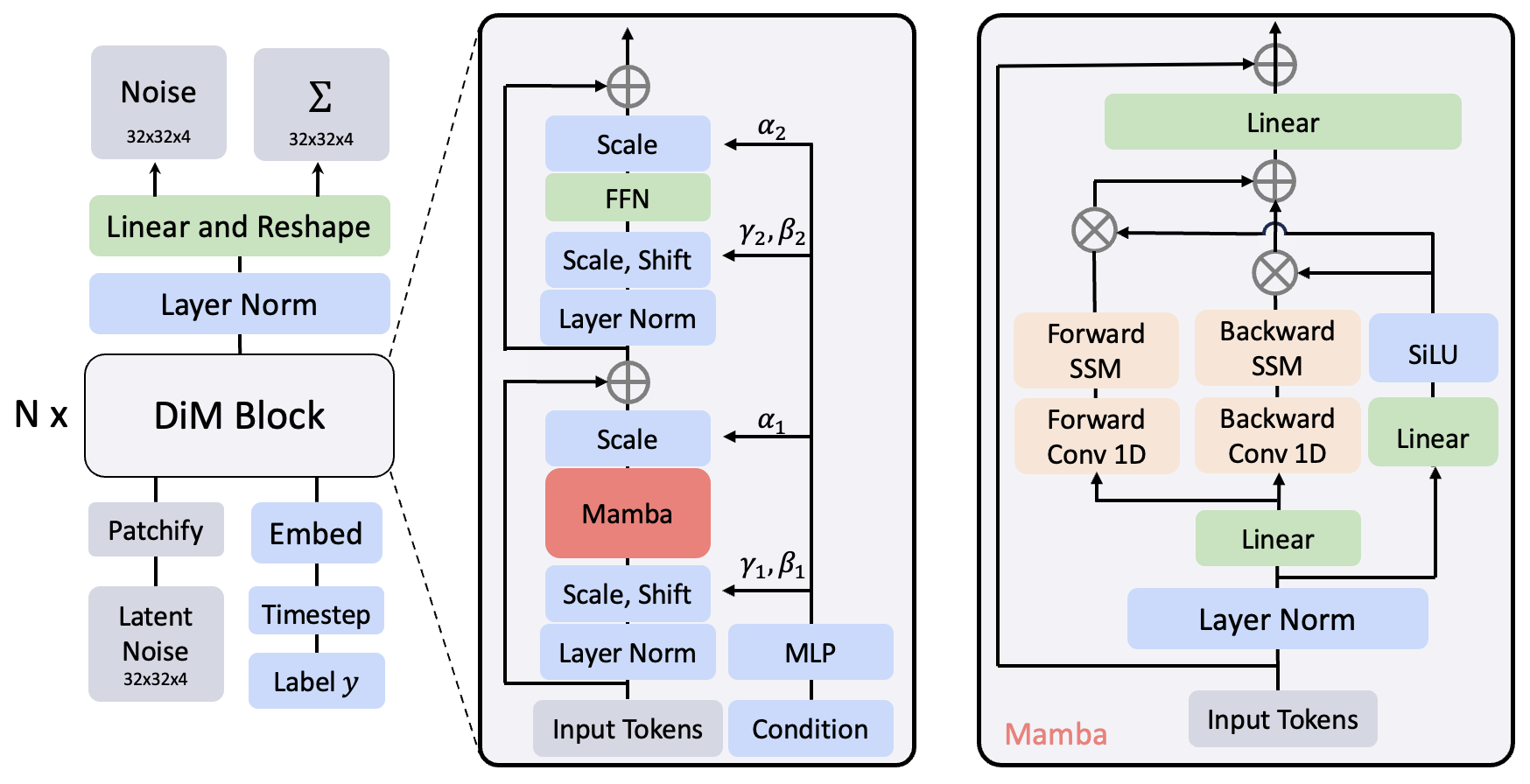}
\caption{{\bf Illustration of the proposed Diffusion Mamba~(DiM) for image generation.}
The pre-trained VAE encoder from Stable Diffusion takes images as input, and a patchification operator is used to generate token-level patch embeddings.
Then, multiple DiM blocks based on Mamba with bidirectional state space models (SSMs) extract representations from all input tokens.
Finally, a linear layer is used to predict the noise in the latent space.
}
\label{fig: main_img}
\end{figure*}

\section{Method}

Our objective is to learn a diffusion-based mamba architecture capable of generating high-fidelity images and videos. 
We introduce the Diffusion Mamba (DiM) architecture that innovatively applies the denoising process of denoising diffusion probabilistic models (DDPMs) to latent image embeddings, as illustrated in Figure~\ref{fig: main_img}. 
The DiM architecture consists of two primary modules: the Diffusion Mamba framework, discussed in Section~\ref{sec:diffmam}, and the DiM block, detailed in Section~\ref{sec:dim}.

\subsection{Preliminaries}

In this section, we first describe the problem setup and notations and then revisit DDPMs and diffusion transformers for image/video generation and SSMs for long sequence modeling.

\noindent\textbf{Problem Setup and Notations.}
We consider a set $\mathcal{S} = \{\mathbf{x}_i\}_{i=1}^S$ of 2D images categorized into $M$ classes. 
Each image $\mathbf{x}_i$ is represented as a tensor in $\mathbb{R}^{H\times W\times 3}$, indicating its height, width, and RGB channels. 
The images are associated with class labels $\{y_i\}^M_{i=1}$, where each $y_i$ indicates the presence of the ground-truth category. 
During training, we use these class labels to facilitate classifier-free guidance in generating conditioned images.

\noindent\textbf{Revisit DDPMs.} 
The image generation problem, the state-of-the-art work~\cite{Rombach2022highresolution} based on denoising diffusion probabilistic models (DDPMs) define a forward noising process that gradually applies noise to latent variable $\mathbf{z}_0$ as $q(\mathbf{z}_t|\mathbf{z}_{t-1}) = \mathcal{N}(\mathbf{z}_t; \sqrt{1-\beta_t}\mathbf{z}_{t-1}, \beta_t\mathbf{I})$, where $\beta_t$ is a Gaussian noise value between $0$ and $1$.
In particular, the denoising process produces a series of shape variables with decreasing levels of noise, denoted as $\mathbf{z}_T, \mathbf{z}_{T-1}, ..., \mathbf{z}_0$, where $\mathbf{z}_T$ is sampled from a Gaussian prior and $\mathbf{z}_0$ is the final output.
With the reparameterization trick, we can have $\mathbf{z}_t = \sqrt{\bar{\alpha}_t}\mathbf{z}_0 + \sqrt{1-\bar{\alpha}_t}\boldsymbol{\epsilon}$, where $\boldsymbol{\epsilon}\sim \mathcal{N}(\mathbf{0},\mathbf{I})$, $\alpha_t = 1-\beta_t$, and $\bar{\alpha}_t = \prod_{i=1}^t \alpha_i$.
For the reverse process, diffusion models are trained to learn a denoising network $\boldsymbol{\theta}$ for inverting forward process corruption as $p_{\boldsymbol{\theta}}(\mathbf{z}_{t-1}|\mathbf{z}_t) = \mathcal{N}(\mathbf{z}_{t-1}; \boldsymbol{\mu}_{\boldsymbol{\theta}}(\mathbf{z}_{t},t), \sigma_t^2\mathbf{I})$.
The training objective is to maximize a variational lower bound of the negative log data likelihood that involves all of $\mathbf{z}_0, ..., \mathbf{z}_T$ as
\begin{equation}
\begin{aligned}
    \mathcal{L} = -p_{\boldsymbol{\theta}}(\mathbf{z}_{0}|\mathbf{z}_1) + \sum_{t} \mathcal{D}_{\text{KL}}(q(\mathbf{z}_{t-1}|\mathbf{z}_{t}, \mathbf{z}_{0})||p_{\boldsymbol{\theta}}(\mathbf{z}_{t-1}|\mathbf{z}_t)))
\end{aligned}
\end{equation}
where $\mathcal{D}_{\text{KL}}(\cdot||\cdot)$ denotes the KL divergence measuring the distance between two distributions.
Since both $p_{\boldsymbol{\theta}}(\mathbf{z}_{t-1}|\mathbf{z}_t))$ and $q(\mathbf{z}_{t-1}|\mathbf{z}_{t}, \mathbf{z}_{0})$ are Gaussians, we can reparameterize $\boldsymbol{\mu}_{\boldsymbol{\theta}}(\mathbf{z}_{t},t)$ to predict the noise $\boldsymbol{\epsilon}_{\boldsymbol{\theta}}(\mathbf{z}_{t},t)$.
In the end, the training objective can be reduced to a simple mean-squared loss between the model output $\boldsymbol{\epsilon}_{\boldsymbol{\theta}}(\mathbf{z}_{t},t)$ and the ground truth Gaussian noise $\boldsymbol{\epsilon}$ as:
$\mathcal{L}_{\text{simple}} = \|\boldsymbol{\epsilon}-\boldsymbol{\epsilon}_{\boldsymbol{\theta}}(\mathbf{z}_{t},t)\|^2$.
After $p_{\boldsymbol{\theta}}(\mathbf{z}_{t-1}|\mathbf{z}_t))$ is trained, new latent variable can be generated by progressively sampling $\mathbf{z}_{t-1}\sim p_{\boldsymbol{\theta}}(\mathbf{z}_{t-1}|\mathbf{z}_t))$ by using the reparameterization trick with initialization of $\mathbf{z}_{T}\sim \mathcal{N}(\mathbf{0},\mathbf{I})$.
DDPMs describe a forward process where a latent variable $\mathbf{z}_0$ is gradually noised to generate a sequence of increasingly noisy states, culminating in a Gaussian distribution. The reverse process involves a trained neural network that incrementally denoises these states to reconstruct the original data. Our model modifies this process to work efficiently with high-resolution images by integrating the Mamba architecture's ability to handle long sequences with reduced complexity.

\noindent\textbf{Revisit Diffusion Transformer (DiT).}
To generate high-fidelity 2D images, DiT~\cite{Peebles2022DiT} proposed to train latent diffusion models (LDMs) with Transformers as the backbone, consisting of two training models.
They first extract the latent code $\mathbf{z}$ from an image sample $\mathbf{x}$ using an autoencoder with an encoder $f_{\text{enc}}(\cdot)$ and a decoder $f_{\text{dec}}(\cdot)$, that is, $\mathbf{z} = f_{\text{enc}}(\mathbf{x})$. 
The decoder is used to reconstruct the image sample $\hat{\mathbf{x}}$ from the latent code $\mathbf{z}$, \textit{i.e.},  $\hat{\mathbf{x}} = f_{\text{dec}}(\mathbf{z})$.
Based on latent codes $\mathbf{z}$, a latent diffusion transformer with multiple designed blocks is trained with time embedding $\mathbf{t}$ and class embedding $\mathbf{c}$, where a self-attention and a feed-forward module are involved in each block.
Note that they apply patchification on latent code $\mathbf{z}$ to extract a sequence of patch embeddings and depatchification operators are used to predict the denoised latent code $\mathbf{z}$. 
DiTs utilized latent embeddings of images, generated through an encoder-decoder structure, to model the diffusion process in a latent space. The DiT enhances the fidelity of generated images by employing transformers that handle these latent embeddings effectively. In DiM, we adapt these principles to work synergistically with Mamba's efficient sequential processing capabilities.

\noindent\textbf{Revisit State Space Models (SSMs).}
State Space Models (SSMs)~\cite{gu2023mamba,gu2022efficiently} are pivotal in our approach due to their efficiency in handling sequences. 
Inspired by continuous systems, Mamba~\cite{gu2023mamba} models map a 1-D function or sequence $x(t) \in \mathbb{R}$ to $y(t) \in \mathbb{R}$, mediated through a hidden state $h(t) \in \mathbb{R}^\mathtt{N}$. The state dynamics are governed by the system matrices $\mathbf{A} \in \mathbb{R}^{\mathtt{N} \times \mathtt{N}}$, $\mathbf{B} \in \mathbb{R}^{\mathtt{N} \times 1}$, and $\mathbf{C} \in \mathbb{R}^{1 \times \mathtt{N}}$:
\begin{equation}
h'(t) = \mathbf{A}h(t) + \mathbf{B}x(t), \quad y(t) = \mathbf{C}h(t).
\end{equation}
For practical applications, Mamba~\cite{gu2023mamba} employs discrete approximations of these continuous systems, facilitated by a transformation parameter $\mathbf{\Delta}$. 
The continuous-to-discrete transformation commonly uses zero-order hold techniques:
\begin{equation}
\mathbf{\overline{A}} = \exp(\mathbf{\Delta} \mathbf{A}), \quad \mathbf{\overline{B}} = (\mathbf{\Delta} \mathbf{A})^{-1} (\exp(\mathbf{\Delta} \mathbf{A}) - \mathbf{I}) \mathbf{\Delta} \mathbf{B}.
\end{equation}
With these transformations, the system equations are discretized, allowing us to compute outputs at specific time steps:
\begin{equation}
h_t = \mathbf{\overline{A}}h_{t-1} + \mathbf{\overline{B}}x_{t}, \quad y_t = \mathbf{C}h_t.
\end{equation}
Finally, the output is computed using a structured global convolution, which processes the sequence through a convolutional kernel derived from the state transition matrices:
\begin{equation}
\mathbf{\overline{K}} = (\mathbf{C}\mathbf{\overline{B}}, \mathbf{C}\mathbf{\overline{A}}\mathbf{\overline{B}}, \dots, \mathbf{C}\mathbf{\overline{A}}^{L-1}\mathbf{\overline{B}}), \quad \mathbf{y} = \mathbf{z} * \mathbf{\overline{K}},
\end{equation}
where $L$ denotes the length of the input sequence $\mathbf{z}$, and $\overline{\mathbf{K}}$ represents a structured convolutional kernel, facilitating efficient sequence processing.

\subsection{Diffusion Mamba}\label{sec:diffmam}

Our Diffusion Mamba (DiM) module adapts the Mamba architecture to handle 2D image data efficiently. This module leverages the linear computational complexity of the Mamba framework, enabling the scaling of image and video frame generation to higher resolutions with reduced computational overhead.
The traditional Mamba model is reconfigured for handling 2D latent code from an image sample $\mathbf{x}$ using an autoencoder with an encoder $f_{\text{enc}}(\cdot)$. 
This involves transforming an image latent $\mathbf{z} \in \mathbb{R}^{H_z \times W_z \times C}$ into a series of flattened 2D patches. 
Each patch $\mathbf{z}_p$ is a vector in $\mathbb{R}^{L \times (P^2 \cdot C)}$, where $(H_z, W_z)$ represents the dimensions of the input latent code, $C$ is the number of channels, and $P$ is the patch size. These patches are linearly projected to a dimensionality of $D$ and combined with position embeddings. A class token is also attached to represent the entire image.
The token-level patch embeddings are processed through the multiple layers of the DiM encoder to produce outputs, iteratively enhancing each patch's feature representation.

\subsection{DiM Block}\label{sec:dim}

The DiM (Diffusion Mamba) block is a key innovation that extends bidirectional sequence modeling to visual tasks by integrating state space modeling with diffusion processes. 
Each block processes input sequences in both forward and backward directions, applying transformations that utilize SSM principles to manage spatial dependencies effectively.
This design ensures that the DiM architecture can generate detailed and coherent images by synthesizing global image features dynamically.

\begin{wraptable}{R}{0.35\textwidth}
\vspace{-12pt}
	\renewcommand\tabcolsep{6.0pt}
    \renewcommand{\arraystretch}{1.1}
	\centering
 \caption{{\bf Detailed configurations of DiM Models.} All models for the Small (S), Base (B), Large (L) and XLarge (XL) settings have comparable parameters to DiT~\cite{Peebles2022DiT} counterparts.}
 \label{tab: model_cfg}
	\scalebox{0.8}{
		\begin{tabular}{l|cc}
		\toprule
Method  & Layers $N$ & Hidden Size $d$\\
  \midrule
DiM-S  & 16         & 384    \\
DiM-B  & 16         & 768    \\
DiM-L  & 32         & 1024   \\
DiM-XL & 36         & 1152  \\
\bottomrule
			\end{tabular}}
\vspace{-0.5em}
\end{wraptable}

Originally, the Mamba architecture was tailored for 1-D sequences and lacked the capability to handle the spatial complexities inherent in vision tasks. 
The DiM block addresses this limitation by incorporating bidirectional sequence modeling optimized for image data, enhancing the system’s ability to comprehend and reconstruct spatial information. 
A schematic paradigm of the DiM block's functionality is illustrated in Figure~\ref{fig: main_img}.
Specifically, the process within each DiM block begins with the normalization of the input token sequence. 
This sequence is then linearly projected to create the state vector $\mathbf{z}$. Subsequently, the vector $\mathbf{z}$ undergoes bidirectional processing. 
For each direction, a 1-D convolution modifies $\mathbf{z}$, creating intermediate states $\mathbf{z}'$, which are then used to derive the transformation parameters $\mathbf{B}$, $\mathbf{C}$, and $\mathbf{\Delta}$.
The outputs for both forward and backward directions are computed using the transformed SSM parameters and subsequently combined to form the output sequence, ensuring effective integration of information from both directions.

The scalability of our DiM blocks is a critical feature, allowing adaptation to various patch sizes and model dimensions. This flexibility enables the architecture to process images ranging in complexity from small to extra-large, accommodating different levels of detail and resolution requirements.
Specifically, it can flexibly accommodate patch dimensions of 2, 4, 8, and model complexity ranging from Small (S), Base (B), Large (L), and XLarge (XL), similar to DiT~\cite{Peebles2022DiT}. 
The detailed configurations are shown in Table~\ref{tab: model_cfg}.

\subsection{Efficiency Analysis}

Our Diffusion Mamba (DiM) architecture has been meticulously designed to optimize both computational and memory efficiency, crucial for scaling up image and video generation tasks to high resolutions. In this section, we compare the efficiency of our DiM model with existing models like the Diffusion Transformer (DiT)~\cite{Peebles2022DiT} and DiffuSSM~\cite{yan2023diffusion}, demonstrating DiM's superior handling of large-scale data without sacrificing performance due to computational or memory constraints.

\textbf{I/O \& Memory Efficiency:} The DiM architecture reduces the need for intensive data transfers between storage and processing units, a common bottleneck in large-scale image processing. By efficiently managing data flow within the system, DiM minimizes read/write cycles, which is vital for speeding up the generation process.
DiM also uses advanced data structuring and caching strategies to minimize memory overhead. This allows the system to handle larger batches of data or higher-resolution images within the constraints of standard hardware.

\textbf{Computation Efficiency:} Perhaps most critically, the computational design of DiM significantly reduces the number of operations required to generate an image compared to other state-of-the-art models. This is quantified as follows:
\begin{itemize}
    \item DiT~\cite{Peebles2022DiT}: $4LD^2 + 2L^2D$
    \item DiffuSSM~\cite{yan2023diffusion}: $7.5LD^2$
    \item DiM (ours): $3L(2D)N + L(2D)N = 8NLD, N=16$
\end{itemize}
Here, $L$ represents the sequence length (which can be correlated with image dimensions), $D$ represents the feature dimensionality, and $N$ is a scaling factor specific to DiM, reflecting the integration of state space modeling techniques that allow bidirectional processing with enhanced efficiency. 
Importantly, the computation cost for DiM is linear with respect to both $L$ and $D$, which underscores its suitability for handling very high-resolution images or long video sequences without exponential increases in computational demand.

\subsection{Scaling to Video Generation}

In extending the Diffusion Mamba (DiM) architecture to video generation, we tackle the complexities associated with maintaining temporal coherence and managing the high resolution of video frames. 
The inherent modular design of the DiM architecture facilitates its adaptation to the dynamic and temporally structured nature of video content, enabling the efficient generation of high-quality video.
Video data introduces an additional temporal dimension that must be seamlessly integrated with spatial processing to ensure the continuity and coherence of generated frames. Our approach involves transforming a video's latent representation, denoted as $\mathbf{z} \in \mathbb{R}^{T\times H_z \times W_z \times C}$, where $T$ represents the time dimension (number of frames), $H_z$ and $W_z$ are the height and width of the latent space, and $C$ is the number of channels.
Each frame's latent space is processed into a series of flattened 2D patches. Specifically, we map $\mathbf{z}$ into: $\mathbf{z}\in\mathbb{R}^{T \times L \times (P^2 \cdot C)}$, where each patch $\mathbf{z}_p$ is a vector in $\mathbb{R}^{T \times L \times (P^2 \cdot C)}$, $L$ denotes the total number of patches per frame, and $P$ represents the size of each patch.

To effectively model the temporal dynamics alongside spatial features, the DiM architecture applies bidirectional sequence processing not only across the spatial dimensions of each frame but also across the temporal sequence of frames. This ensures that each generated frame is informed by its predecessors and successors, thus preserving temporal continuity.
The sequential processing across time is achieved by extending the bidirectional capabilities of our state space models, allowing them to capture dependencies not just within frames but also between successive frames. This dual focus on spatial and temporal features is crucial for generating video content that is not only high in visual quality but also smooth and consistent in motion.
Through this approach, the DiM architecture leverages its computational efficiency and scalability to handle the increased demands of video generation, making it capable of producing high-resolution video content efficiently, without compromising on the temporal fidelity essential for realistic and engaging media playback.

\section{Experiments}

\subsection{Experimental Setup}

\noindent\textbf{Datasets.}
For image generation, we utilize the ImageNet~\cite{imagenet_cvpr09} dataset, which is standard in evaluating generative models due to its complexity and variety. For video generation, we employ the UCF-101~\cite{soomro2012ucf101} dataset, a widely recognized benchmark in the video domain. From UCF-101, we extract 16-frame video clips using a specific sampling interval, with each frame resized to 256x256 resolution to standardize the input for our model.

\noindent\textbf{Evaluation Metrics.}
For image generation, we follow DiT~\cite{Peebles2022DiT} and measure performance using the Frechet Inception Distance (FID)~\cite{heusel2018gans}, Inception Score~\cite{salimans2016improved}, sFID~\cite{nash2021generating}, and Precision/Recall~\cite{kynkäänniemi2019improved} metrics. Specifically, we calculate FID-50K using 250 DDPM sampling steps to ensure robustness in our evaluations.
For video generation, we assess our model using the Frechet Video Distance (FVD), adhering to the evaluation guidelines introduced by StyleGAN-V~\cite{ivan2022styleganv}. We compute the FVD scores by analyzing 2,048 video clips, each comprising 16 frames, to ensure statistical significance and comparability.

\noindent\textbf{Implementation.}
Our implementation is based on PyTorch~\cite{paszke2019PyTorch}.
For image generation, our model follows DiT~\cite{Peebles2022DiT} using the AdamW optimizer with a constant learning rate of 1×1e-4, no weight decay, and a batch size of 256. 
The only data augmentation technique employed is horizontal flipping. 
We maintain an exponential moving average (EMA) of DiT weights over training with a decay of 0.9999, and all reported results use the EMA model.
For video generation, our approach follows the Latte~\cite{ma2024latte} model specifications. 
We utilize AdamW optimizer settings similar to those for image generation and maintain an EMA of Latte weights with a decay rate of 0.9999. Additionally, we use the pre-trained variational autoencoder from Stable Diffusion~\cite{Rombach2022highresolution} 1.4 to handle the latent representations effectively.

\begin{table}[t]
	\renewcommand\tabcolsep{6.0pt}
    \renewcommand{\arraystretch}{1.1}
	\centering
 \caption{{\bf Comparison results on all metrics of our \colorbox{red!10}{DiM-B/4} and DiT-B/4 models across 400K training steps on ImageNet 256x256.} }
 \label{tab: exp_256_step}
	\scalebox{0.9}{
		\begin{tabular}{l|ccccccc}
		\toprule
Method  & Steps & CFG & FID ($\downarrow$) & sFID ($\downarrow$) & IS ($\uparrow$) & Precision ($\uparrow$) & Recall ($\uparrow$) \\
  \midrule
DiT-B/4 & 100K  & 1.5 & 79.45     & \bf 14.62      & 17.41    & 0.31            & \bf 0.41         \\
\rowcolor{red!10}
DiM-B/4 & 100K  & 1.5 & \bf 74.69     & 16.13      & \bf 19.38    & \bf 0.34            & 0.39         \\ \hline
DiT-B/4 & 200K  & 1.5 & 59.02     & \bf 11.26      & 25.01    & 0.39            & 0.49         \\
\rowcolor{red!10}
DiM-B/4 & 200K  & 1.5 & \bf 53.69     & 12.14      & \bf 29.42    & \bf 0.44            & \bf 0.50          \\ \hline
DiT-B/4 & 300K  & 1.5 & 49.89     & \bf 10.38      & 30.69    & 0.44            & \bf 0.52         \\
\rowcolor{red!10}
DiM-B/4 & 300K  & 1.5 & \bf 43.57     & 10.83      & \bf 37.47    & \bf 0.49            & 0.51         \\ \hline
DiT-B/4 & 400K  & 1.5 & 44.43     & \bf 9.95       & 35.03    & 0.47            & \bf 0.53         \\
\rowcolor{red!10}
DiM-B/4 & 400K  & 1.5 & \bf 37.62     & 10.20       & \bf 43.93    & \bf 0.52            & \bf 0.53        \\ 
\bottomrule
			\end{tabular}}
   \vspace{-1em}
\end{table}

\subsection{Comparison to prior work}

In this work, we propose a novel and effective diffusion transformer for image and video generation.
In order to validate the effectiveness of the proposed DiM, we comprehensively
compare it to previous image and video generation baselines~\cite{Peebles2022DiT,tulyakov2017mocogan,yan2021videogpt,tian2021a,yu2022generating,ivan2022styleganv,yu2023video,shen2023mostganv,he2022lvdm,ma2024latte}.

\noindent\textbf{Image generation across all training steps.}
In this analysis, we evaluate the performance of the DiM architecture throughout the training process to understand its learning dynamics and stability. 
We report the performance across all training steps in Table~\ref{tab: exp_256_step}, showcasing the scalability and efficiency improvements over traditional models.
By tracking metrics such as FID-50K at various checkpoints, we observe how quickly the model converges to a high-quality synthesis compared to DiT~\cite{Peebles2022DiT}.

\begin{wraptable}{R}{0.4\textwidth}
\vspace{-10pt}
\centering
\caption{{\bf Comparison results with video generation models on UCF101 dataset}. The FVD values are reported and ``IMG'' denotes video-image joint training.}
\label{tab: exp_video}
\begin{tabular}{lc}
\toprule
Method            & UCF101 \\ \midrule
MoCoGAN~\cite{tulyakov2017mocogan}           & 2886.9 \\
VideoGPT~\cite{yan2021videogpt}          & 2880.6 \\
MoCoGAN-HD~\cite{tian2021a}        & 1729.6 \\
DIGAN~\cite{yu2022generating}             & 1630.2 \\
StyleGAN-V~\cite{ivan2022styleganv}        & 1431.0 \\
PVDM~\cite{yu2023video}              & 1141.9 \\
MoStGAN-V~\cite{shen2023mostganv}         & 1380.3  \\
LVDM~\cite{he2022lvdm}              &  372.00   \\ \hline
Latte~\cite{ma2024latte}            &  477.97 \\
Latte~\cite{ma2024latte}+IMG        &  333.61  \\ 
\rowcolor{red!10}
DiM (ours) & \bf 358.75 \\
\rowcolor{red!10}
DiM+IMG (ours) & \bf 206.83 \\  \bottomrule
\end{tabular}
\vspace{-1.0em}
\end{wraptable}

\noindent\textbf{Image generation across all model sizes.}
Scaling the model size from small (S) to extra large (XL) allows us to assess how well the DiM architecture scales with increased computational resources. 
This segment of the analysis focuses on comparing the performance across various model sizes using the FID-50K metric. 
We evaluate the DiM architecture across various model sizes (S/4, B/4, L/4, XL/4) and compare it to the DiT model on the FID-50k metric. 
The results, illustrated in Figure~\ref{fig: title_image}, highlight the superior performance of our model across different scales.

\noindent\textbf{Video generation on UCF-101 dataset.}
For video generation, we use the UCF-101 dataset to test the temporal coherence and visual quality of generated video clips. The Frechet Video Distance (FVD) metric is employed to quantitatively measure these aspects by comparing the statistical distribution of generated videos against real videos from the dataset. 
The capabilities of DiM in video generation are quantified in Table~\ref{tab: exp_video}, emphasizing its ability to produce high-fidelity and diverse video content efficiently.
This analysis not only demonstrates the effectiveness of the DiM in capturing and reproducing complex temporal dynamics but also benchmarks its performance against existing video generation models.

\begin{figure*}[t]
\centering
\includegraphics[width=0.99\linewidth]{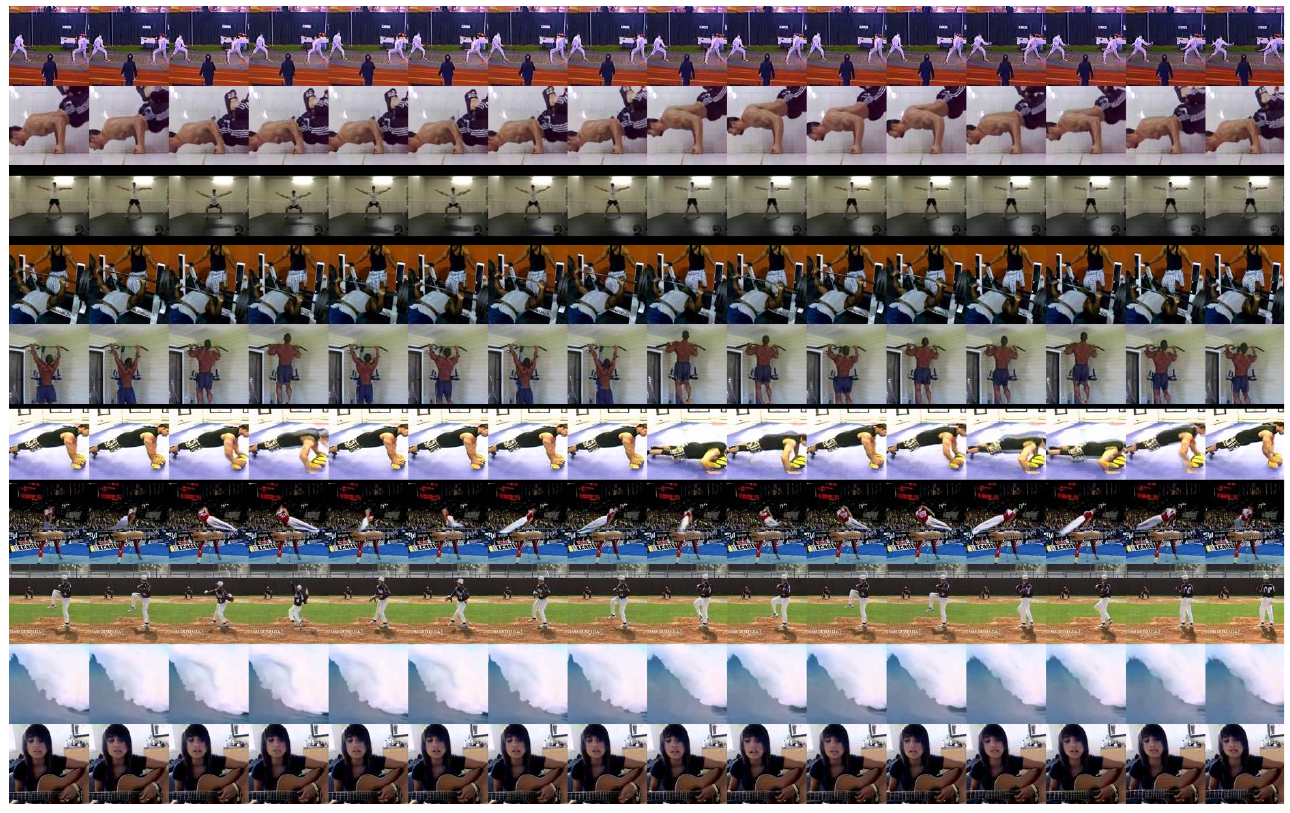}
\caption{{\bf Qualitative visualization of video generation on UCF-101 datasets.} Our DiM can operate the denoising process from videos to generate high-fidelity and diverse video clips.
}
\label{fig: exp_vis}
\end{figure*}

\noindent\textbf{Qualitative visualizations.}
Besides quantitative metrics, qualitative assessments are crucial for generative models. 
These qualitative results in Figure~\ref{fig: exp_vis} showcase the effectiveness of applying a plain diffusion mamba architecture to operate the denoising process from videos for generating high-fidelity and diverse videos.
These visual examples also demonstrate both the strengths and limitations of the DiM, such as handling complex scenes or motion blur in videos.

\begin{table}[t]
	\renewcommand\tabcolsep{6.0pt}
    \renewcommand{\arraystretch}{1.1}
	\centering
 \caption{{\bf Ablation results on Gflops of \colorbox{red!10}{DiM-XL/2} models versus DiT-XL/2 models across diverse image resolutions.} 
 All Gflops are calculated by thop package for a fair comparison.}
 \label{tab: ab_flops}
	\scalebox{0.98}{
		\begin{tabular}{l|cc}
		\toprule
Method  & Image Resolution & Model Gflops ($\downarrow$)  \\
  \midrule
DiT-XL/2 & 256x256        & 343.28                  \\
\rowcolor{red!10}
DiM-XL/2 & 256x256        & \bf 294.58                  \\ \hline
DiT-XL/2 & 512x512        & 1371.08                 \\
\rowcolor{red!10}
DiM-XL/2 & 512x512        & \bf 1175.7                  \\ \hline
DiT-XL/2 & 1024x1024       & 5482.28                 \\
\rowcolor{red!10}
DiM-XL/2 & 1024x1024       & \bf 4700.17                 \\ \hline
DiT-XL/2 & 2048x2048       & OOM \\
\rowcolor{red!10}
DiM-XL/2 & 2048x2048       & \bf 18798.04  \\       
\bottomrule
			\end{tabular}}
   \vspace{-1em}
\end{table}

\subsection{Experimental analysis}

In this section, we performed ablation studies to demonstrate the efficiency of introducing mamba on image generation.
We also conducted extensive experiments to explore the scalability of patch size and model size, and the influence of classifier-free guidance.

\noindent\textbf{Efficiency of Mamba on Gflops.}
We quantitatively evaluate the computational efficiency of our model by measuring the Gflops required for generating images with and without the Mamba architecture. This metric provides insight into the computational cost reductions achieved by our method. Using the \textit{thop} package~\footnote{\url{https://github.com/Lyken17/pytorch-OpCounter}}, we calculate the total number of floating-point operations (Flops). The results, detailed in Table~\ref{tab: ab_flops}, highlight the significant reduction in Gflops when employing the Mamba architecture, demonstrating its potential for enabling more efficient generative processes.

\begin{table}[t]
	\renewcommand\tabcolsep{6.0pt}
    \renewcommand{\arraystretch}{1.1}
	\centering
 \caption{{\bf Ablation results on Gflops and image generation metrics of all DiM-B models with 400K training steps on ImageNet 256x256.} All images are generated with CFG=1.5.}
 \label{tab: ab_patch}
	\scalebox{0.95}{
		\begin{tabular}{l|ccccccc}
		\toprule
Method  & Params (M) & Gflops ($\downarrow$) & FID ($\downarrow$) & sFID ($\downarrow$) & IS ($\uparrow$) & Precision ($\uparrow$) & Recall ($\uparrow$) \\
  \midrule
DiM-B/8 & 134.82 & \bf 3.83  & 96.10      & 27.27      & 15.20     & 0.27            & 0.30          \\
DiM-B/4 & 134.37 & 14.71 & 37.62     & 10.20       & 43.93    & 0.52            & \bf 0.53         \\
DiM-B/2 & 134.26 & 58.23 & \bf 15.39     & \bf 5.76       & \bf 88.25    & \bf 0.69            & 0.52        \\
\bottomrule
			\end{tabular}}
\end{table}

\begin{table}[t]
	\renewcommand\tabcolsep{6.0pt}
    \renewcommand{\arraystretch}{1.1}
	\centering
 \caption{{\bf Ablation results on Gflops and image generation metrics of all DiM models with 400K training steps on ImageNet 256x256.} All models are trained with a patch size of 4, and images are generated with CFG=1.5.}
 \label{tab: ab_model}
	\scalebox{0.95}{
		\begin{tabular}{l|ccccccc}
		\toprule
Method  & Params (M) & Gflops ($\downarrow$) & FID ($\downarrow$) & sFID ($\downarrow$) & IS ($\uparrow$) & Precision ($\uparrow$) & Recall ($\uparrow$) \\
  \midrule
DiM-S/4  & 33.71  & \bf 3.69  & 71.03     & 17.65      & 21.54    & 0.36            & 0.41         \\
DiM-B/4  & 134.37 & 14.71 & 37.62     & 10.2       & 43.93    & 0.52            & 0.53         \\
DiM-L/4  & 473.73 & 52.22 & 20.67     & 7.44       & 75.14    & 0.63            & 0.53         \\
DiM-XL/4 & 673.82 & 74.33 & \bf 17.26     & \bf 6.93       & \bf 86.62    & \bf 0.66            & \bf 0.53        \\
\bottomrule
			\end{tabular}}
\end{table}

\begin{table}[!ht]
	\renewcommand\tabcolsep{6.0pt}
    \renewcommand{\arraystretch}{1.1}
	\centering
 \caption{{\bf Impact of Classifier-free Guidance (CFG) on all metrics of our \colorbox{red!10}{DiM-B/4} and DiT-B/4 models across 400K training steps on ImageNet 256x256.} CFG=1.0 denotes that images are generated without Classifier-free Guidance.}
 \label{tab: ab_cfg}
	\scalebox{0.98}{
		\begin{tabular}{l|ccccccc}
		\toprule
Method  & Steps & CFG & FID ($\downarrow$) & sFID ($\downarrow$) & IS ($\uparrow$) & Precision ($\uparrow$) & Recall ($\uparrow$) \\
  \midrule
DiT-B/4 & 400K  & 1.0   & 67.42     & \bf 12.62      & 20.62    & 0.37            & 0.55         \\
\rowcolor{red!10}
DiM-B/4 & 400K  & 1.0   & \bf 61.99     & 13.23      & \bf 23.67    & \bf 0.40             & \bf 0.56         \\
DiT-B/4 & 400K  & 1.5 & 44.43     & \bf 9.95       & 35.03    & 0.47            & \bf 0.53         \\
\rowcolor{red!10}
DiM-B/4 & 400K  & 1.5 & \bf 37.62     & 10.20       & \bf 43.93    & \bf 0.52            & \bf 0.53        \\
\bottomrule
			\end{tabular}}
\end{table}

\noindent\textbf{Scaling Patch size.}
The size of patches into which the images are decomposed represents a critical hyperparameter in our architecture. We explore how varying patch sizes affect the image quality and computational demands of the DiM model. A systematic analysis across different patch dimensions provides insights into the optimal configurations for balancing performance with computational efficiency. The findings in Table~\ref{tab: ab_patch} help identify the most effective patch size for enhancing the model's efficiency and output fidelity.

\noindent\textbf{Scaling Model size.}
This part of our analysis focuses on understanding how the DiM scales with changes in model size, from small to extra-large configurations. By evaluating the impact of model size on image quality (measured by FID) and computational requirements, we can determine the scalability of our architecture. 
The results reported in Table~\ref{tab: ab_model}, illustrate the trade-offs between increased computational resources and improvements in image generation quality.

\noindent\textbf{Influence of Classifier-free Guidance.}
Classifier-free guidance has been touted as a method to improve the conditional generation quality by reducing the dependency on specific classifier outputs during training. We examine the impact of this technique on the performance of DiM in generating conditionally guided images. This analysis in Table~\ref{tab: ab_cfg}, assesses whether removing direct classifier dependencies in the generation process improves the versatility and quality of the generated images.

\section{Conclusion}

In this work, we present DiM, a novel approach that incorporates the efficiency and scalability of the Mamba architecture into the realm of image and video generation. 
By eschewing traditional attention mechanisms in favor of state space models, DiM offers a significant reduction in computational complexity and operational overhead, making it particularly well-suited for generating high-resolution media.
Our experimental results demonstrate that DiM not only reduces the computational demands but also scales effectively with changes in input size and model dimensions. 
The ablation studies further validate the model's efficiency, showing significant improvements in processing time and resource utilization without compromising the quality of the generated images and videos.
Moreover, by exploring the impact of patch size, model size, and classifier-free guidance, we have established that DiM can adapt to various configurations and requirements, proving its versatility across different scenarios. The qualitative and quantitative assessments confirm that DiM can produce high-fidelity and diverse outputs, setting a new standard for future generative models.

\noindent\textbf{Limitation.}
Despite the significant advancements introduced by the DiM architecture, some limitations warrant further exploration. 
The adaptation of DiM to video generation, although promising, has not been exhaustively tested across a wide range of video types and qualities. 
Its performance in scenarios with highly dynamic content or low visibility remains to be fully assessed. Furthermore, the current implementation might not capture long-term dependencies effectively over extended video sequences, which is critical for applications like long-form content generation.

\noindent\textbf{Broader Impact.}
The DiM architecture presents numerous opportunities for positive impact across various sectors. In the media and entertainment industry, DiM can be used to generate high-quality content efficiently, potentially reducing costs and increasing the accessibility of media production. In areas such as education and training, the ability to create detailed visual content can enhance learning materials and simulations.

\clearpage

\bibliography{reference}
\bibliographystyle{unsrt}




\end{document}